\documentclass{article}
\usepackage{amssymb}

\usepackage[preprint]{corl_2025} 

\usepackage{graphicx}
\usepackage[table,dvipsnames]{xcolor}         
\usepackage{algorithm, algorithmicx, algpseudocode}
\usepackage{multirow}
\usepackage{bbm}
\usepackage{amsmath, amssymb}
\usepackage[utf8]{inputenc} 
\usepackage[T1]{fontenc}    
\usepackage{multicol}
\usepackage{hyperref}       
\usepackage{url}            
\usepackage{booktabs}       
\usepackage{amsfonts}       
\usepackage{amsmath, amssymb, amsfonts}
\usepackage{caption}
\usepackage{nicefrac}       
\usepackage{microtype}      
\usepackage{subcaption}
\usepackage{cleveref}
\usepackage[disable]{todonotes}
\usepackage{amsthm}
\usepackage{marginnote}
\newtheorem{theorem}{Theorem}
\newtheorem{prop}[theorem]{Prop}

\title{Group Policy Gradient}

\author{
  Junhua Chen$^\dagger$\\
  University of Cambridge, 
  United Kingdom\\
  \texttt{jc2318@cam.ac.uk} \\
  \And
  Zixi Zhang$^\dagger$\\
  University of Cambridge, 
  United Kingdom\\
  \texttt{zz458@cam.ac.uk} \\
  \And
  Hantao Zhong\\
  University of Cambridge, 
  United Kingdom\\
  \texttt{hz445@cam.ac.uk} \\
  \And
  Rika Antonova\\
  University of Cambridge, 
  United Kingdom\\
  \texttt{ra702@cam.ac.uk} \\
}

\begin{document}

\maketitle

\def\thefootnote{$\dagger$}\footnotetext{These authors contributed equally to this work}\def\thefootnote{\arabic{footnote}}

\begin{abstract}
We introduce Group Policy Gradient (GPG), a family of \emph{critic-free} policy-gradient estimators for general MDPs. Inspired by the success of GRPO’s approach in Reinforcement Learning from Human Feedback (RLHF), GPG replaces a learned value function with a group-based Monte Carlo advantage estimator, removing the memory, compute, and hyperparameter costs of training a critic while preserving PPO’s clipped-objective structure. We prove the consistency of the GPG estimator, analyze the bias-variance tradeoffs, and demonstrate empirically that GPG matches or outperforms PPO on standard 
benchmarks. GPG makes better use of parallel simulations, which, together with its critic-free design, results in more efficient use of computational resources than PPO. 
\end{abstract}

\section{Introduction}

Reinforcement learning (RL) trains agents to maximize cumulative rewards through interaction with an environment~\citep{RL-an-introduction}. Policy gradient methods~\citep{vpg} combined with deep networks excel in domains from game playing~\citep{Silver2016AlphaGo} to continuous control~\citep{schulman2018highdimensionalcontinuouscontrolusing} and generative modeling~\citep{uehara2024understandingreinforcementlearningbasedfinetuning}. Proximal Policy Optimization (PPO)~\citep{ppo}, with its clipped objective for stable updates, is now a default choice in deep RL and a core method in Reinforcement Learning from Human Feedback (RLHF)~\citep{rlhf} for fine-tuning language models from human preferences.

Until recently, PPO dominated RLHF, leveraging a learned value function (critic) to reduce variance in gradient estimates and improve learning efficiency. However, critics add computational and memory overhead and are prone to approximation errors. Group Relative Policy Optimization (GRPO)~\citep{deepseekmath} addresses these issues by estimating advantages through a group-based Monte Carlo approach, removing the need for a value network. This critic-free design enabled large language models to match or exceed prior RLHF performance, particularly on mathematical reasoning, while substantially reducing memory and computational cost~\citep{deepseekr1}.

Building on the strengths and resource-saving benefits of critic-free, group-based policy gradients, as well as their success in RLHF, we extend these methods to the broader realm of general RL. Our contributions are as follows:

\begin{itemize}
    \item We generalize the GRPO framework and introduce a new critic-free policy gradient algorithm for general Markov Decision Processes (MDPs), which we call Group Policy Gradient (GPG). Like GRPO, GPG modifies only the advantage estimation step while preserving the core structure and benefits of PPO.
    \item We prove the consistency of our resulting policy gradient estimator under mild assumptions, showing that it converges in the large-group-size limit.
    \item We empirically evaluate GPG on various Gymnasium environments, validating its effectiveness with standard 
    RL benchmarks. We perform ablation studies on design choices and discuss the practical trade-offs of our method.
\end{itemize}

\begin{figure}
    \centering
    \includegraphics[width=\linewidth]{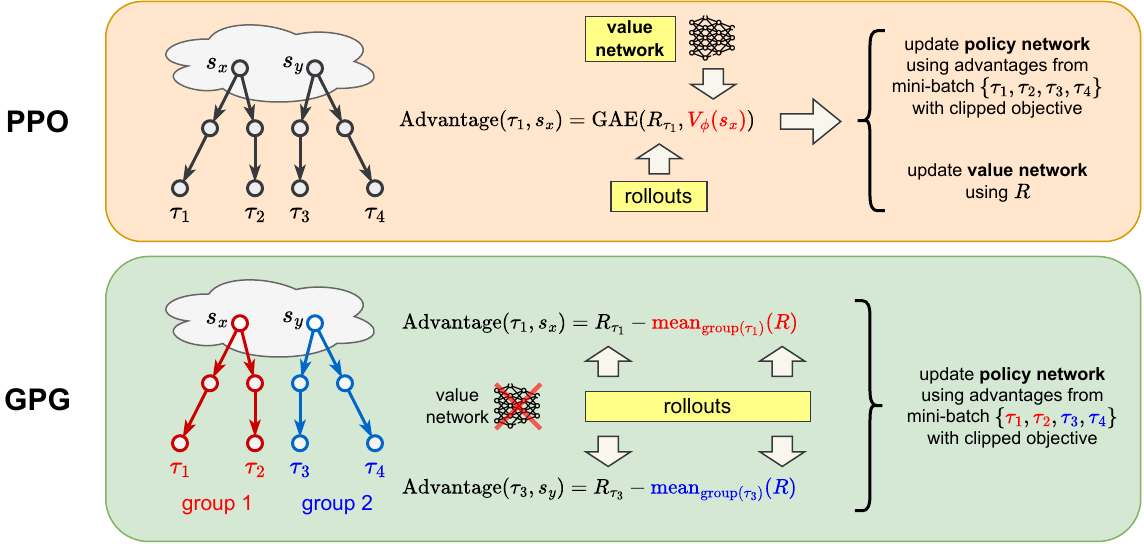}
    \caption{PPO (top) estimates the advantage function using Generalized Advantage Estimation (GAE) with the aid of a learnt value function. In contrast, GPG (bottom) utilizes group-averaged rewards to reduce policy gradient variance. GPG avoids learning a value function and makes greater use of the information in parallel simulations, thereby making better use of computational resources.}

    \label{fig:grpo-vs-ppo-adv-est}

\end{figure}

\section{Background}

\subsection{Notations and Conventions}
\label{app:not-conv}

We largely follow the convention of the Generalized Advantage Estimation paper~\citep{schulman2018highdimensionalcontinuouscontrolusing}: We consider an undiscounted formulation of the policy optimization problem.
The initial state  $s_0$ is sampled from the distribution $\rho_0$.
A trajectory $\tau=(s_0, a_0, s_1, a_1, \dots)$ is generated by sampling actions according to the policy $a_t \sim \pi(a_t | s_t)$ and sampling the states according to the dynamics $s_{t+1} \sim P(s_{t+1} | s_t, a_t)$, until a terminal (absorbing) state is reached.
A reward $r_t = r(s_t,a_t,s_{t+1})$ is received at each timestep.
The goal is to maximize the expected total reward $\sum_{t=0}^{\infty} r_t$, which is assumed to be finite for all policies.
Note that we are not using a discount as part of the problem specification; it will appear below as an algorithm parameter that adjusts a bias-variance tradeoff.
But the discounted problem (maximizing $\sum_{t=0}^{\infty} \gamma^t r_t$) can be handled as an instance of the undiscounted problem in which we absorb the discount factor into the reward function, making it time-dependent.

We also utilize standard definitions for Value function and the $Q$ function:
\begin{itemize}
\item $V^\pi(s) = \mathbb{E}_\pi \left[ \sum_{t=0}^{\infty} r_t \mid S_0 = s \right]$
\item $Q^\pi(s, a) = \mathbb{E}_\pi \left[ \sum_{t=0}^{\infty} r_t \mid S_0 = s, A_0 = a \right]$
\end{itemize}

For a sampled trajectory $\tau$, we also define the time $t$ $\gamma$-discounted total return $R^\gamma_{t}(\tau)=\sum_{s\geq t}\gamma^{s-t}r_s$





\subsection{Advantage Estimation and Baselines}\label{section:advantage-estimate}
A crucial result, first observed in~\citep{vpg} and generalized in~\citep{schulman2018highdimensionalcontinuouscontrolusing}, in policy-gradient algorithms, describes a family of policy-gradient estimators and is given by the following proposition; See~\citep{schulman2018highdimensionalcontinuouscontrolusing} for the proof.
\begin{prop}\label{Prop:AdvantageEstimation}
For independent trajectories $\tau_{1:N}$ sampled under policy $\pi_\theta$, any estimator of the form
\begin{equation}
    \sum_{i=1}^N \sum_{t=1}^T \nabla_\theta\log \pi_\theta(a_t^{(i)}|s_t^{(i)})\hat A_t^{(i)}
\end{equation}
where $\hat A_t^{(i)}=q(s_{t:\infty}^{(i)},a_{t:\infty}^{(i)}, r_{t:\infty}^{(i)})-b(s_{1:t}^{(i)},a_{1:(t-1)}^{(i)})$, where $q$ is any function $q$ that satisfying $\mathbb{E}[q(s_{t:\infty}^{(i)},a_{t:\infty}^{(i)})|s_t,a_t]=Q(s_t,a_t)$ and $b$ is \textit{any} function leads to an unbiased estimator for $\nabla_\theta \eta$. 
\end{prop}

The quantity $\hat A$ is often referred to by \textit{advantage estimator}, and $b$ by \textit{baseline}. A judicious choice of $q$ and $b$ leads to lower variance estimates, with $b=V^{\pi_\theta}$ the on-policy value function known to give near-optimal~\citep{schulman2018highdimensionalcontinuouscontrolusing} estimates from a variance perspective. However, with $V^{\pi_\theta}$ intractable, it is common in algorithms such as PPO and TRPO to learn an approximate value function $\hat V_\phi$ (usually represented as a neural network, trained concurrently using gradient descent) alongside $\pi_\theta$ to reduce variance in advantage estimation. Well-tuned advantage estimators like Generalized Advantage Estimation~\citep{schulman2018highdimensionalcontinuouscontrolusing} are critical to the success of algorithms like PPO.

\subsection{Proximal Policy Optimization}
\label{section:ppobackground}

Among policy gradient algorithms, PPO~\citep{ppo} improves training stability by constraining policy updates, using a relatively cheap clipped surrogate objective:
\begin{equation}
\label{PPOLoss}
\mathcal{L}=\mathbb{E}_{\tau\sim\pi_{\theta_{\mathrm{old}}}}\left[\min\left\{
\frac{\pi_\theta(a_t|s_t)}{\pi_{\theta_{\mathrm{old}}}(a_t|s_t)}\hat A_t,\  \mathrm{clip}\left(\frac{\pi_\theta(a_t|s_t)}{\pi_{\theta_{\mathrm{old}}}(a_t|s_t)}, 1-\epsilon, 1+\epsilon\right)\hat A_t
\right\}\right]
\end{equation}
where \(\hat A_t\) is commonly computed using truncated GAE:
\begin{equation}\label{eq:GAE}
\hat A_t=\sum_{s=t}^T(\gamma\lambda)^{s-t}\delta_s,\qquad
\delta_t=r_t+\gamma V(s_{t+1})-V(s_t).    
\end{equation}

PPO is simple and delivers excellent performance on a wide range of tasks, but requires learning a value network, which is often costly and introduces additional complexity ~\citep{ppo,OpenAI_PPO,juliani2024studyplasticitylossonpolicy}.

\subsection{Group Relative Policy Optimization}\label{section:GRPObackground}

GRPO~\citep{deepseekmath} replaces the learned critic in PPO with a \emph{group}-based, locally estimated baseline. In the RLHF-style \textit{Outcome-Supervision} setting (one scalar reward \(r^{(i)}\) per trajectory), GRPO sets advantages for each trajectory in a batch of size \(N\) to the group-normalized reward:
\begin{equation}\label{eq:GRPOAdvantages}
\hat A_t^{(i)}=\frac{r^{(i)}-\operatorname{mean}(\mathbf r)}{\operatorname{std}(\mathbf r)},\quad t=1,\dots,T,
\end{equation}
(or the analogous cumulative form for \textit{Process Supervision}). Equivalently, GRPO applies REINFORCE with a locally estimated baseline, avoiding a learned value network. This critic-free design drives efficiency gains in LLM fine-tuning~\citep{deepseekmath,deepseekr1}, but has so far been explored mainly in RLHF settings rather than general MDPs.

\subsection{Other Group-based RL Algorithms}
Following GRPO’s success, several RL algorithms adopting group-based advantage estimation have emerged. \citet{sane2025hybridgrouprelativepolicy} propose Hybrid-GRPO, an adaptation of group estimation to general RL environments, though without clear experimental validation. The recently introduced REINFORCE++~\citep{hu2025reinforcesimpleefficientapproach} is effectively PPO with advantage normalization and $\lambda_{\text{GAE}}=1$, equivalent to replacing GAE with discounted returns and thus eliminating the critic. While effective in RLHF, it has not been evaluated in general RL settings. GiGPO~\citep{gigpo} extends GRPO by normalizing advantages across episodes and states from different trajectories within a group, showing strong results on LLM agent benchmarks but not on broader RL tasks.

\section{Group Policy Gradient}
\label{sec:method}
In this section, we first introduce a method that computes advantage estimates for each state within a group of sampled trajectories (\Cref{alg:GPGAdvantage}). These estimates are then used by the PPO algorithm to update the policy, forming the GPG Algorithm (\Cref{alg:GPG}). We then explore the design space of GPG methods and present a theoretical result on Group Policy Gradient Estimation.

\subsection{The GPG Method}

\begin{figure}
\centering
\begin{minipage}{.48\textwidth}
    \begin{algorithm}[H]
    \caption{GPG Update}
    \label{alg:GPG}
    \begin{algorithmic}[1]
    \State Initialize policy $\pi_{\theta}$
    \For{each iteration}
        \State Set $\pi_{\theta_\text{old}}\leftarrow \pi_\theta$
        \State Collect trajectories \textcolor{red}{group} $\tau_{1:N}$ under $\pi_\theta$
        \State \textcolor{red}{Compute advantage estimates $\hat A_t^n$ c.f \Cref{alg:GPGAdvantage}}
        \For{several epochs}
            \State Optimize PPO loss $\mathcal{L}_{PPO}$ by gradient descent on mini-batch estimates
        \EndFor
        \State Update policy $\pi_{\theta}$ via gradient ascent                               
    \EndFor
    \end{algorithmic}
    \end{algorithm}
\end{minipage}
\hfill
\begin{minipage}{.48\textwidth}
    \begin{algorithm}[H]
    \caption{GPG Advantage Estimation}
    \label{alg:GPGAdvantage}
    \begin{algorithmic}[1]
    \Require Trajectories $\tau_{1:N}$, Binning function $f$
    \State Compute returns $R_t^n$ for all $t$ and $n=1\ldots N$
    \State Initialize Empty Bins $B$
    \For{$n=1\ldots N$}
        \For{$t=1\ldots$}
        \State If $f(s_t^n,t)\neq f(s_i^n,i),\ i=1\ldots t-1$, insert $R_t^n$ to $B[f(s_t^{(n)},t)]$ 
        \EndFor
    \EndFor
    \State Set $\hat A_t^{(n)}=R_t^{(n)}-\text{mean}(B[f(s_t^{(n)},t)])$ for all $t,n$
    \end{algorithmic}
    \end{algorithm}
\end{minipage}
\end{figure}


GPG departs from PPO and GRPO only in how it estimates advantages, making it simple to implement and scale. It generalizes GRPO’s group-based variance-reduction technique, using a broader formulation that applies to any RL setting. Unless noted otherwise, we denote $R_t^{(i)}=R^\gamma_t(\tau_i)$ the time $t$ discounted returns for the $i$th trajectory of the group. 

For GPG, we first introduce the concept of a \textit{binning} function $f:\mathcal{S}\to \mathcal{B}$ where $\mathcal{B}$ is a countable set of bins and $\mathcal{S}$ is the \textit{timestep-aware} state space. We use this to divide the set of states into a set of bins, with $f(s)$ being the bin state $s$ is in. This in turn lets us define the \textit{bin value function} $b(s)=\mathbb{E}_{s'\sim\pi_\theta|f(s')=f(s)}[V(s)]$ which corresponds to a state-likelihood weighted average of state-value functions of the bin each state is in. Inspired by Prop~\ref{Prop:AdvantageEstimation}, we will use estimates of $b(s)$ as our policy-gradient baseline. 

Formally, given independent trajectories $\tau_{1:N}$ sampled for a \textit{group}, we estimate advantages using \begin{equation}\label{eq:GPGadvtanges}
    \hat A_t^{(i)}=R_t^{(i)}-\widehat b_N(s_t^{(i)})
\end{equation} where $\widehat b_N(s)$ is an estimate of the (on-policy) value function at $s$ from the group trajectories.  We take the estimated state-value $\hat b_N(s)$ to be the average discounted return from the bin of $s$ i.e \begin{equation}
    \hat b_N(s)=\text{mean}(\{R_t^{(i)}|(i,t):f(s_t^i)=f(s) \text{ and $t$ is first visit in $\tau_i$ to a state in bin $f(s)$}\})
\end{equation}
We illustrate this advantage calculation in \Cref{alg:GPG} and \Cref{alg:GPGAdvantage}. In our paper, we consider several different possibilities of $f$, based on time and spatial partitioning of the states. We will see in the experimental section that for the purposes of policy gradient estimation, there is a need to strike a balance in the bin granularity, with both too fine or too coarse a bin size being counter-productive to agent learning. Here, we highlight 4 bin functions to give examples:
\begin{itemize}
    \item $f(s, t)=0$, where only 1 bin is present
    \item $f(s, t)=t$, where an average baseline is computed for each timestep
    \item $f(s, t)=\epsilon\cdot \text{Round}(s/\epsilon)$ the discretization of space into $\epsilon$-sized packets, for continuous state spaces in $\mathbb{R}^d$. 
    \item $f(s,t)=s$ for discrete state spaces, where each state is its own bin
\end{itemize}

\paragraph{Relation of GPG to GRPO, PPO and Sampling Methods:} Our framework generalizes GRPO with Outcome Supervision. To see this, note that Outcome supervision corresponds to using the trivial binning function $f(s)=0,\ \forall s\in \mathcal{S}$, where the same mean reward (equal to return when only a terminal reward is given) is subtracted as the baseline for all states, as well as PPO advantage normalization~\citep{shengyi2022the37implementation} afterwards. Moreover, using $f(s)=0$ with advantage normalization also corresponds to the REINFORCE++ Algorithm~\citep{hu2025reinforcesimpleefficientapproach}. Similarly, process supervision can be interpreted as subtracting a group-estimated baseline from the $R_t^n$s, then normalizing. The subtracted baseline depends only on the timestep used, which bears similarity to the choice $f(s,t)=t$. In general, our algorithm is identical to PPO and GRPO, with the sole difference being in our advantage-estimation mechanism. 

We further note that the idea of using group-estimated quantities for estimation variance reduction has its roots in Monte-Carlo literature, such as with control variates~\citep{goodman2005variance}. The trade-off between a learned and group-estimated control variate also features in GFlownet and Variational Inference literature, such as through comparing the Vargrad and Trajectory Balance Losses~\citep{malkin2023trajectorybalanceimprovedcredit, richter2020vargradlowvariancegradientestimator}.

\paragraph{Remark:} The form we choose to take for the binning function is not the most general. In fact, as Prop~\ref{Prop:AdvantageEstimation} suggests, the binning function can be modified suitably to take the whole \textit{history} up to time $t$ as input. However, for simplicity, we only study binning functions of the current state.

\subsection{Theoretical Guarantees of GPG}

Many Policy Gradient Algorithms, ranging from the simple REINFORCE~\citep{vpg} method to PPO itself, has the property that if only one gradient step is taken over each batch of data, then as the batch size tends to infinity the gradient update converges in probability to the true policy gradient $\nabla_\theta \eta(\theta)$. Here, we prove the corresponding statement for GPG. We omit technical conditions and only present a proof sketch here, deferring the full proof and details to the \Cref{appendix:proof}. 
\begin{prop}\label{Prop:Consistency} Consider a MDP environment without discounting and with fixed duration $T$ steps\footnote{This constraint, merely for ease of notation and analysis, is easily relaxable}. For a group of $N$ iid trajectories $\tau_{1:N}$ sampled from $\pi_{\theta_\text{old}}$, the GPG Policy-Gradient estimator  is \begin{equation}\label{eq:GPGGradient}
    \nabla_\theta \mathcal{L}_N=\nabla_\theta \left[ \frac{1}{N}\sum_{n=1}^N \sum_{t=1}^T  \min(\hat A_t^n\frac{\pi_\theta(a_t^n|s_t^n)}{\pi_{\theta_\text{old}}(a_t^n|s_t^n)}, \hat A_t^n\text{clip}(\frac{\pi_\theta(a_t^n|s_t^n)}{\pi_{\theta_\text{old}}(a_t^n|s_t^n)},1-\epsilon,1+\epsilon))\right]
\end{equation} where as before $A_t^{n}=R_t^{(i)}-\widehat b_N(s_t^{n})$ for some countably-valued binning function $f$. Then we have 
\paragraph{(i)} In the case of one update per group i.e $\pi_\theta = \pi_{\theta_\text{old}}$, we have \begin{equation}\label{eq:gradientSimplifies}
\nabla_\theta\mathcal{L}_N=\nabla_\theta \left[ \frac{1}{N}\sum_{n=1}^N \sum_{t=1}^T  \hat A_t^n\frac{\pi_\theta(a_t^n|s_t^n)}{\pi_{\theta_\text{old}}(a_t^n|s_t^n)} \right]= \frac{1}{N}\sum_{n=1}^N \sum_{t=1}^T  \hat A_t^n\nabla_\theta \log \pi_\theta(a_t^n|s_t^n)
\end{equation}

\paragraph{(ii)} Moreover in the one update case, assuming some regularity assumptions (see \Cref{appendix:proof}), the GPG gradient estimator is a consistent estimator (in group size $N$) of the policy gradient i.e \begin{equation}
    \nabla_\theta \mathcal{L}_N\longrightarrow^\mathbb{P} \nabla_\theta\eta(\tau)
\end{equation}
\end{prop}

\paragraph{Proof Sketch:} \textbf{(i)} is the same as a well-known result for PPO and GRPO (c.f~\citep{OpenAI_PPO} or~\citep{deepseekmath}). For \textbf{(ii)}, a direct application of the strong law of large numbers (SLLN) is not possible as the $\hat A_t^{n}$ is not independent for different trajectories. Nevertheless, the proof makes heavy use of SLLN, and follows the intuition that for all $s$, $\hat b_N(s)\to^{\text{almost sure}} b(s)$ as $N\to \infty$, for the bin-value function $b(s)$. This is a valid baseline function as per Prop~\ref{Prop:AdvantageEstimation}. Some care must be taken when there is an infinite number of bins, but we can show that for sufficiently large group sizes, we can visit \textit{most} of the states (in a likelihood weighted sense) sufficiently often to get good estimates of $b(s)$ for each state.  
\paragraph{Corollary:} We can also show that the GRPO Policy-Gradient estimator is a consistent estimator of the normalized policy gradient $\nabla_\theta\eta(\tau)/\text{std}(R_1)$. To see that, note that the GRPO gradient estimator is $\pmb{g}_{N}=\frac{\nabla\mathcal{L}_N}{\hat \sigma_N}$ where $\hat \sigma_N=\sqrt{\frac{1}{N}\sum_{i=1}^N(R_1^{{i}})^2-\left(\frac{1}{N}\sum_{i=1}^N(R_1^{{i}})\right)^2}$ is the usual standard deviation estimator of the total returns. By standard results we have that $\hat \sigma_N\to^\mathbb{P}\text{std}(R_1)$, and so by Slutsky's Lemma we have the desired result $\pmb{g}_N\to^\mathbb{P} \nabla_\theta\eta(\tau)/\text{std}(R_1)$.

\section{Experiments}

\begin{figure}
    \centering
    \includegraphics[width=1.0\linewidth]{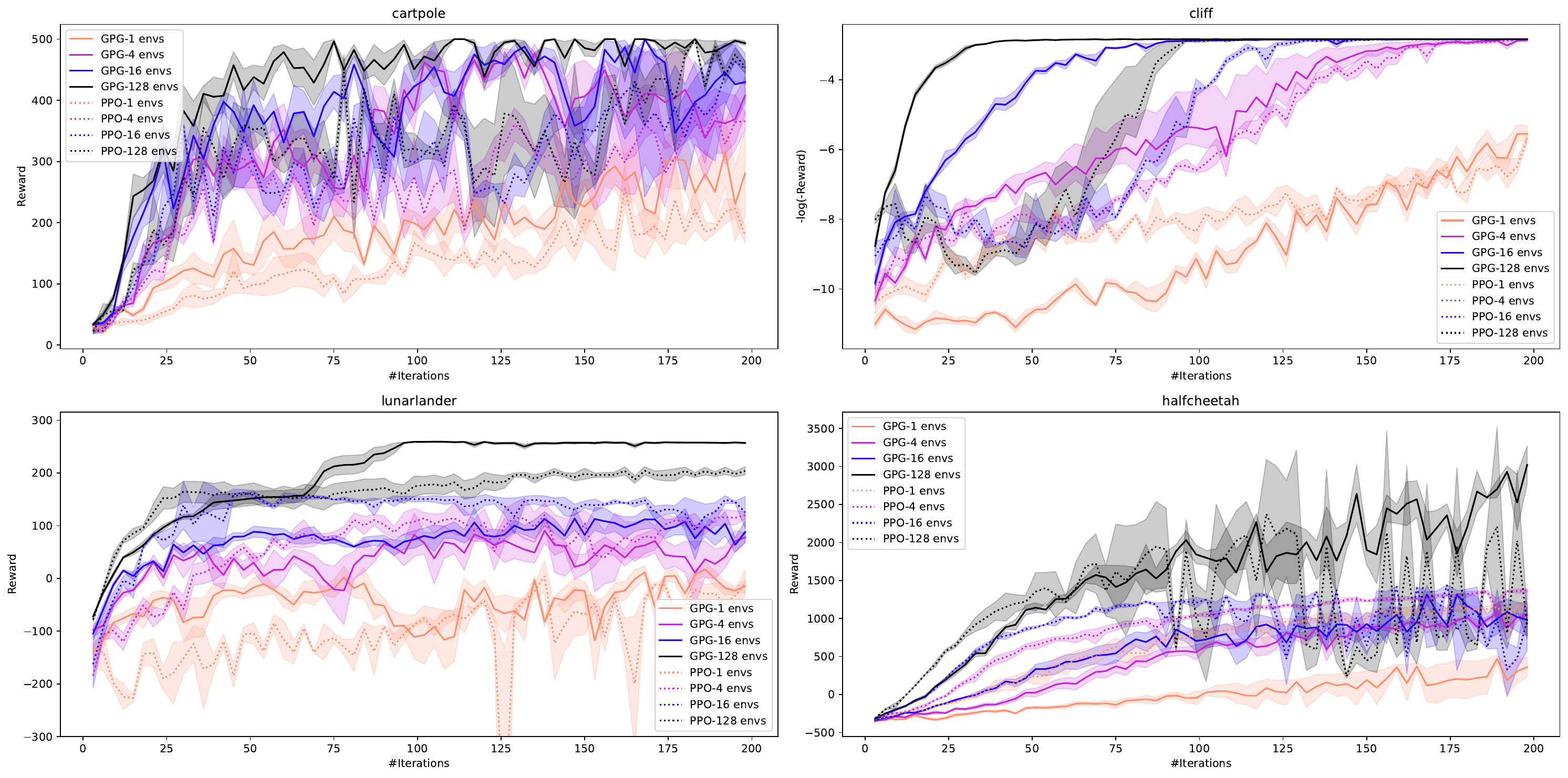}
    \caption{Average episodic reward of GPG and PPO for different numbers of parallel environments. For clarity, we plot on a logarithmic scale for the CliffWalker environment. Given a large number of parallel environments, GPG dominates on all tasks.} 
    \label{fig:main-result-iter}
\end{figure}

We conduct a comprehensive comparison between PPO and GPG across a suite of reinforcement learning environments from the OpenAI Gymnasium library~\citep{towers2024gymnasium}. Additionally, we present ablation studies to further analyze performance differences.

\subsection{Experimental Setup}

We build GPG by modifying the advantage-estimation component of the reference CleanRL PPO implementation~\citep{huang2021cleanrlhighqualitysinglefileimplementations}, and otherwise reuse standard PPO hyperparameters when applicable. In GPG each rollout from a vectorized environment forms a \emph{group}, so the nominal group size equals the number of parallel environments (automatic resets can increase the effective size). Because GPG can be sensitive to group size, we sweep the number of parallel environments and report evaluation reward curves for PPO and GPG (5 evaluation seeds, 4 training seeds per configuration), training each run for 200 iterations. All GPG experiments use the time-binning \(f(s,t)=t\). We evaluate on four Gymnasium tasks (CartPole, CliffWalker, LunarLander, HalfCheetah) using default environment parameters; further implementation and hyperparameter details are in \Cref{section:environmentdetails}.

\subsection{Results}
\label{sec:experiments-main-results}

\begin{table}[t]
    \centering
    \caption{Rewards for GPG and PPO for various numbers of parallel environments (equivalent to group size for GPG) after 200 Iterations. GPG with group size 128 exhibits dominant performance on all benchmarks.}
    \label{tab:main-results}
    \begin{tabular}{ll|llll}
        \toprule
        \# Rollouts          & Algorithm & CartPole                       & CliffWalker                     & Half-Cheetah                     & Lunarlander                    \\ 
        \midrule
        \multirow{2}{*}{1}   & GPG      & $255.73\scriptstyle{\pm44.20}$ & $-261.30\scriptstyle{\pm70.56}$ & $333.70\scriptstyle{\pm94.87}$   & $-18.78\scriptstyle{\pm6.76}$  \\
                             & PPO       & $205.82\scriptstyle{\pm10.05}$ & $-442.93\scriptstyle{\pm73.40}$ & $679.33\scriptstyle{\pm278.55}$  & $-19.58\scriptstyle{\pm12.73}$ \\ 
        \midrule
        \multirow{2}{*}{4}   & GPG      & $388.65\scriptstyle{\pm23.65}$ & $-17.45\scriptstyle{\pm0.11}$   & $1031.65\scriptstyle{\pm24.77}$  & $74.29\scriptstyle{\pm10.33}$  \\
                             & PPO       & $316.85\scriptstyle{\pm31.80}$ & $-17.62\scriptstyle{\pm0.15}$   & $1346.30\scriptstyle{\pm15.33}$  & $117.77\scriptstyle{\pm5.39}$  \\
        \midrule
        \multirow{2}{*}{16}  & GPG      & $428.05\scriptstyle{\pm33.31}$ & $\pmb{-17.00\scriptstyle{\pm0.00}}$   & $1000.05\scriptstyle{\pm130.02}$ & $75.81\scriptstyle{\pm12.26}$  \\
                             & PPO       & $423.17\scriptstyle{\pm22.28}$ & $\pmb{-17.00\scriptstyle{\pm0.00}}$   & $736.45\scriptstyle{\pm149.77}$  & $157.59\scriptstyle{\pm10.03}$ \\
        \midrule
        \multirow{2}{*}{32} & GPG & $481.10\scriptstyle{\pm10.85}$ & $-17.15\scriptstyle{\pm0.09}$ & $1940.86\scriptstyle{\pm142.08}$ & $169.67\scriptstyle{\pm20.79}$ \\
                             & PPO       & $442.80\scriptstyle{\pm15.04}$ & $\pmb{-17.00\scriptstyle{\pm0.00}}$   & $1142.48\scriptstyle{\pm570.77}$ & $157.77\scriptstyle{\pm10.00}$ \\
        \midrule
        \multirow{2}{*}{128} & GPG      & $\pmb{495.45\scriptstyle{\pm2.13}}$  & $\pmb{-17.00\scriptstyle{\pm0.00}}$   & $\pmb{2773.61\scriptstyle{\pm222.93}}$ & $\pmb{257.39\scriptstyle{\pm0.80}}$  \\
                             & PPO       & $474.20\scriptstyle{\pm7.61}$  & $\pmb{-17.00\scriptstyle{\pm0.00}}$   & $1516.90\scriptstyle{\pm305.42}$ & $200.97\scriptstyle{\pm4.42}$  \\
        \bottomrule
    \end{tabular}
\end{table}

Compared to PPO, GPG exhibits strong performance on all tasks. We comment on the following. 
\paragraph{Performance of different methods:} As indicated in \Cref{fig:main-result-iter}, both PPO and GPG generally benefit from larger numbers of parallel environments. However, due to its use of group-based baselines, the improvement from this parallelism is more pronounced for GPG, similar to GRPO and as hinted by the proof of Prop~\ref{Prop:Consistency}. As a result, for all considered tasks, GPG with large group size performs the best, as displayed in \Cref{tab:main-results} or \Cref{fig:main-result-iter}. 

\begin{figure}
    \centering
    \includegraphics[width=1.0\linewidth]{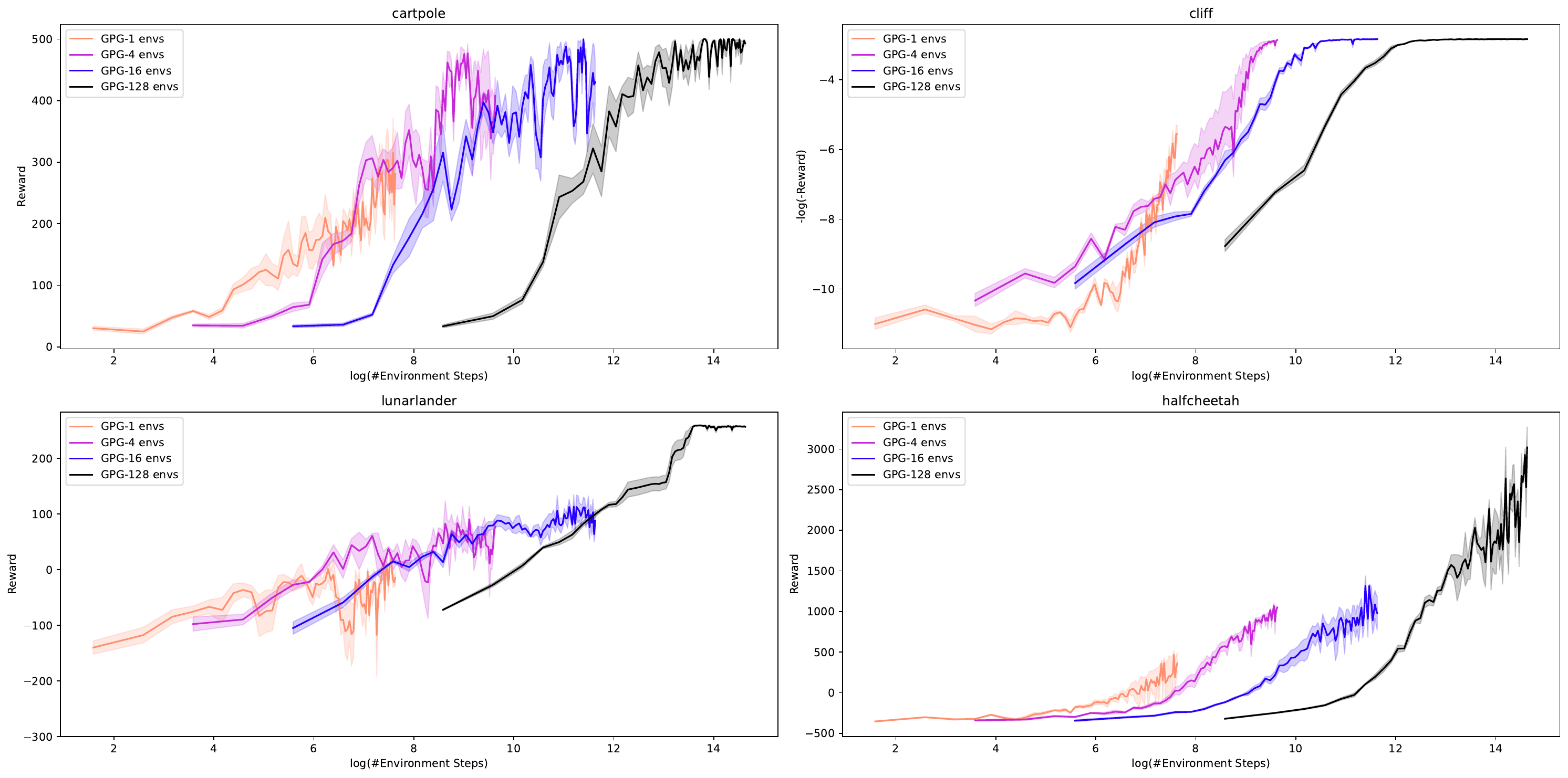}
    \caption{Average episodic rewards for GPG with varying numbers of parallel environments are shown, plotted against the number of evaluated environment steps on a logarithmic scale for clarity. While increasing the number of parallel environments generally reduces sample efficiency, requiring more total environment interactions (but fewer iterations) to reach a given performance threshold, it leads to higher iteration-based performance.}
    \label{fig:main-result-step}
\end{figure}

\paragraph{Sample Efficiency:} 
For a fixed number of parallel environments, GPG matches PPO in sample efficiency. In some tasks, such as LunarLander or Half-Cheetah (\Cref{fig:main-result-iter}), PPO learns slightly faster early on, but GPG reliably catches up, and with large group sizes, often surpasses PPO. As shown in \Cref{fig:main-result-step}, increasing the number of parallel environments reduces sample efficiency: for a fixed budget of environment steps, training with fewer environments over more iterations generally yields better results. This is intuitive: more iterations give the agent more opportunities to refine its policy, and sequential updates are more expressive than an equivalent amount of parallel computation.

However, small environment counts tend to cap final performance below what large group sizes can achieve, as demonstrated in \Cref{tab:main-results}. In other words, more parallel environments improve iteration-based performance at the cost of sample efficiency. In settings where parallel simulation is cheap and abundant, iteration-based performance becomes the more relevant metric, and our experiments show that GPG excels under such conditions.


\subsection{Ablation Studies}

\begin{figure}
    \centering
    \includegraphics[width=\linewidth]{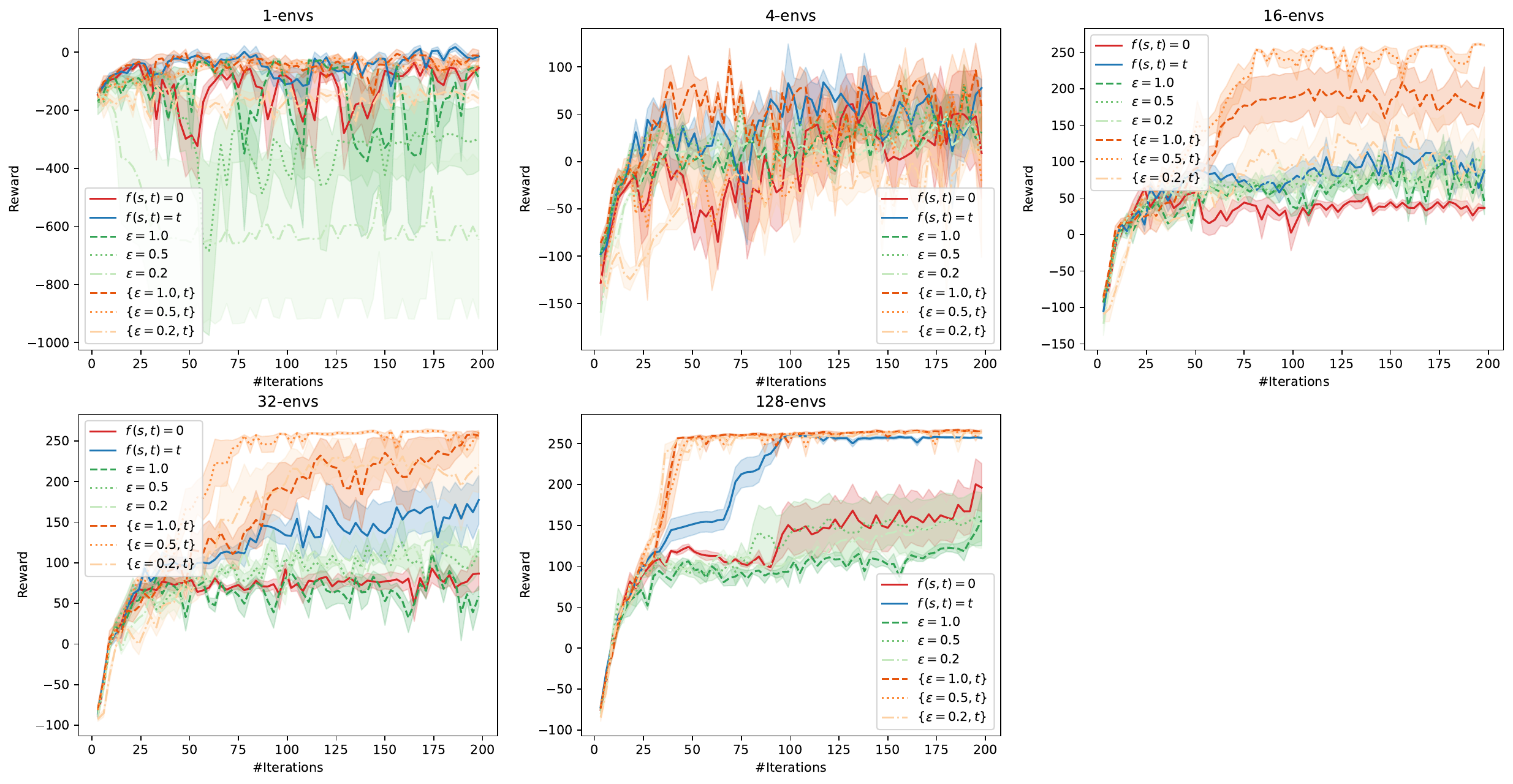}
    \caption{Average episodic reward of GPG on LunarLander with different binning functions.}
    \label{fig:abla-binning}
\end{figure}

We compare four types of binning functions on the LunarLander task:
\begin{enumerate}
    \item \textbf{Time-based binning}: $f(\mathbf s,t)=t$, where the baseline value is averaged over samples in the same timestep.
    \item \textbf{Universal binning}: $f(\mathbf s,t)=0$, where all samples share a single bin, regardless of state or time.
    \item \textbf{Spatial-based binning}: $f(\mathbf s,t) = \epsilon\cdot\mathbf{Round}(\mathbf s/\epsilon)$, which discretizes the continuous state space into bins of size $\epsilon$ along each dimension. To keep this manageable, we test $\epsilon=1.0$, $0.5$, and $0.2$ uniformly across all state dimensions.
    \item \textbf{Spatial-time-based binning}: $f(\mathbf s,t) = \{\epsilon\cdot\mathbf{Round}(\mathbf s/\epsilon), t\}$, combining spatial and time-based binning. States are grouped only if they fall into the same spatial bin and occur at the same timestep.
\end{enumerate}

These choices trade bias and variance: coarser bins reduce bias but keep high variance (like REINFORCE), while very fine bins lower variance at the cost of bias and small-sample issues. \Cref{fig:abla-binning} summarizes results. Universal binning consistently underperforms time-based binning. Spatial binning rarely beats time-based binning except at large environment counts, indicating heterogeneous returns within spatial bins. Spatial-time-based matches time-based performance at small parallelism but improves as the number of environments grows: \(\epsilon=1.0\) works best at 1–4 envs, \(\epsilon=0.5\) at intermediate counts (16–32), and \(\epsilon=0.2\) only matches coarser settings once parallelism is very large (128 envs). This progression underscores how increased environment counts mitigate low-sample issues in finer-grained bins. When more environments are available, the higher trajectory count neutralizes the bias from small sample sizes in each bin. As a result, finer-grained binning becomes advantageous, offering better variance reduction and ultimately leading to faster, more stable convergence. 

\section{Conclusion}
We propose Group Policy Gradient (GPG), a simple generalization of PPO and GRPO that replaces the learned critic with a group-based advantage estimator. We prove consistency of the GPG policy-gradient estimator and provide empirically validated guidelines for key design choices (notably group size and binning). Empirically, we demonstrate that GPG matches or exceeds PPO on a suite of standard 
benchmarks, and analyze its practical tradeoffs and limitations. By replacing the value network with a group-estimated baseline, GPG reduces memory and computational overhead, making it especially attractive when training a critic is costly or unstable, as well as utilizing parallel simulations more efficiently. We hope this study motivates further analysis of variance reduction mechanisms, such as through group baselines, in policy gradient algorithms.

\clearpage

\bibliography{references}

\newpage
\appendix
\section{Formal Statement and Proof of Prop~\ref{Prop:Consistency}}\label{appendix:proof}

\paragraph{Technical Conditions:} Assume the regularity condition that $\sup_{\theta, s,a} \|\nabla_\theta \log \pi_\theta(a|s)\|=C<\infty$ and that all rewards per action are bounded in magnitude by $r_\text{max}<\infty$. Assume further that the policy have nonzero probabilities of visiting states in every bin.

\paragraph{Proof:}
\paragraph{(i):} Same as discussed in~\citep{deepseekmath} A.1.6. When $\pi_\theta=\pi_{\theta_\text{old}}$ the clip and min operations are irrelevant, leading to the first equality. The final equality uses the fact that $\nabla_\theta \log f(\theta)=\frac{\nabla_\theta f(\theta)}{f(\theta)}$. 
\begin{equation}
\begin{aligned}
    \nabla_\theta \mathcal{L}_N&=\nabla_\theta \left[ \frac{1}{N}\sum_{n=1}^N \sum_{t=1}^T  \hat A_t^n\frac{\pi_\theta(a_t^n|s_t^n)}{\pi_{\theta_\text{old}}(a_t^n|s_t^n)} \right]\\&= \frac{1}{N}\sum_{n=1}^N \sum_{t=1}^T  \hat A_t^n\frac{\nabla_\theta \pi_\theta(a_t^n|s_t^n)|_{\theta=\theta_\text{old}}}{\pi_{\theta_\text{old}}(a_t^n|s_t^n)}\\&=\frac{1}{N}\sum_{n=1}^N \sum_{t=1}^T  \hat A_t^n\nabla_\theta \log \pi_\theta(a_t^n|s_t^n)
\end{aligned}
\end{equation}
\paragraph{(ii):} Starting from the simplified form of the gradient in \textbf{(i)}, we have
\begin{equation*}
\begin{aligned}\nabla\mathcal{L}_N &=\frac{1}{N}\left[ \sum_{n=1}^N \sum_{t=1}^T  ( R_t^n+b(s_t^n))\nabla_\theta \log \pi_\theta(a_t^n|s_t^n)+ \sum_{n=1}^N \sum_{t=1}^T  (\hat b_N(s_t^n)-b(s_t^n))\nabla_\theta \log \pi_\theta(a_t^n|s_t^n)\right]\
\end{aligned}
\end{equation*}

where recall $b(s)=\mathbb{E}_{s'\sim\pi_\theta|f(s')=f(s)}[V(s)]=\mathbb{E}_{\tau,s_0'\sim\pi_\theta|f(s'_0)=f(s)}[R(s'_0)]$ is the average (weighted by the policy distribution over states) value function (or equivalently expected forward-return) of a state in the same bin as $s$. Observe that for $n=1\ldots N$, $$\sum_{t=1}^T  ( R_t^n+b(s_t^n))\nabla_\theta \log \pi_\theta(a_t^n|s_t^n)$$ is now independently and identically distributed, and moreover corresponds to the REINFORCE gradient estimator with baseline $b$. As $\nabla_\theta \log \pi_\theta(a_t^n|s_t^n)$ is bounded and $r_t^n$ (and by consequence $R_t^n$ and $b(s_t^n)$) is bounded, the Law of Large Numbers is applicable and thus by Prop~$\ref{Prop:AdvantageEstimation}$ we have 
\begin{equation}
\begin{aligned}
\frac{1}{N}\left[ \sum_{n=1}^N \sum_{t=1}^T  ( R_t^n+b(s_t^n))\nabla_\theta \log \pi_\theta(a_t^n|s_t^n)\right]&\to^\mathbb{P}\mathbb{E}[\sum_{t=1}^T  ( R_t+b(s_t))\nabla_\theta \log \pi_\theta(a_t|s_t)]\\&=\nabla_\theta\eta(\theta)
\end{aligned}
\end{equation}

Recall standard results that if $X_n\to^\mathbb{P} c$ and $Y_n\to^\mathbb{P} d$ then $X_n+Y_n\to^\mathbb{P} c+d$ and $X_nY_n\to^\mathbb{P} cd$. To establish \textbf{(ii)}, it thus suffice to show that\begin{equation}
    \frac{1}{N}\left[ \sum_{n=1}^N \sum_{t=1}^T  (\hat b_N(s_t^n)-b(s_t^n))\nabla_\theta \log \pi_\theta(a_t^n|s_t^n)\right]\to^\mathbb{P} \pmb{0}
\end{equation}

We will show this by checking $N\to \infty$, \begin{equation}K_N=\mathbb{E}_{\tau_{1:N}\sim \pi_\theta}\left[\left\|\frac{1}{N}\sum_{n=1}^N \sum_{t=1}^T  (\hat b_N(s^n_t)-b(s^n_t))\nabla_\theta \log \pi_\theta(a_t^n|s_t^n)\right\|\right]\to 0\end{equation} as convergence in $L^1$ implies convergence in $\mathbb{P}$. The key intuition guiding this part of the proof is that for sufficiently large group sizes, we can visit \textit{most} of the bins (in a visit frequency weighted sense) sufficiently often to get good baseline estimates. By triangle inequality and boundedness, have \begin{equation}
    \begin{aligned} K_N
&\leq\mathbb{E}\left[\frac{1}{N}\sum_{n=1}^N \sum_{t=1}^T  |\hat b_N(s^n_t)-b(s^n_t)|\|\nabla_\theta \log \pi_\theta(a_t^n|s_t^n)\|\right]\\
& \leq\mathbb{E}\left[ \frac{C}{N}\sum_{n=1}^N \sum_{t=1}^T  |\hat b_N(s^n_t)-b(s^n_t)|\right]\end{aligned}
\end{equation}

Now observe that $\sum_{t=1}^T  |\hat b_N(s^n_t)-b(s^n_t)|$ is identically distributed for $n=1\ldots N$ (by symmetry) so by linearity of expectation we obtain 
$$\mathbb{E}\left[ \frac{C}{N}\sum_{n=1}^N \sum_{t=1}^T  |\hat b_N(s^n_t)-b(s^n_t)|\right]=C\mathbb{E}\left[\sum_{t=1}^T  |\hat b_N(s^N_t)-b(s^N_t)|\right]=C\sum_{t=1}^T \mathbb{E}\left[ |\hat b_N(s^N_t)-b(s^N_t)|\right]$$

Next, let $\beta:\mathcal{B}\to\mathbb{R}$ be the mapping from bins to expected-bin value (i.e $\beta(f(s))=b(s)$) and $\hat \beta_N$ to be the empirical bin value function from $\tau_{1:N}$ (i.e such that $\hat \beta_N(f(s))=\hat b_N(s)$). Then we have 
\begin{equation}
\begin{aligned}
C \mathbb{E}\left[\sum_{t=1}^T |\hat b_N(s_t^N)-b(s_t^N)|\right] &=C \mathbb{E}\left[\sum_{t=1}^T |\hat \beta_N(f(s_t^N))-\beta(f(s_t^N))|\left(\sum_{x\in \mathcal{B}}\mathbbm{1}_{f(s_t^N)=x}\right)\right]\\
&=C \mathbb{E}\left[\sum_{x\in \mathcal{B}}\sum_{t=1}^T |\hat \beta_N(x)-\beta(x)|\mathbbm{1}_{f(s_t^N)=x}\right]\\
&=C \sum_{x\in \mathcal{B}}\sum_{t=1}^T\mathbb{E}\left[ |\hat \beta_N(x)-\beta(x)|\mathbbm{1}_{f(s_t^N)=x}\right]\\
&=C \sum_{x\in \mathcal{B}}\sum_{t=1}^T\mathbb{E}\left[ |\hat \beta_N(x)-\beta(x)|\ |\text{bin } x\text{ nonempty}\right]\mathbb{E}[\mathbbm{1}_{f(s_t^N)=x}]\\
\end{aligned}
\end{equation}
where the last equality follows from the fact that $\hat \beta_N(x)$ depends only on rewards which happen after visiting $x$, and where we switch order of summation and expectation by convergence theorems. Letting $\rho(x)=\mathbb{E}[\sum_{t=1}^T \mathbbm{1}_{f(s_t^N)=x}]$, we have 
\begin{equation}
    \begin{aligned}
       C \mathbb{E}\left[\sum_{t=1}^T |\hat b_N(s_t^N)-b(s_t^N)|\right] &= C \sum_{x\in \mathcal{B}}\mathbb{E}\left[ |\hat \beta_N(x)-\beta(x)|\ |\text{bin } x\text{ nonempty}\right]\sum_{t=1}^T\mathbb{E}[\mathbbm{1}_{f(s_t^N)=x}]\\
        &= C\sum_{x\in \mathcal{B}}\rho(x)\mathbb{E}\left[|\hat \beta_N(x)-\beta(x)|\ |\text{bin } x\text{ nonempty}\right]
    \end{aligned}
\end{equation}

It thus remains to show $\sum_{x\in \mathcal{B}} \rho(x)\mathbb{E}_{\tau_{1:N}}\left[|\hat \beta_N(x)-\beta(x)|\ |\text{bin } x\text{ nonempty}\right]\to 0$ as $N\to \infty$. Let $\epsilon>0$ be given in the definition of convergence. We use several facts and observations:
\begin{itemize}
\item As rewards are bounded in magnitude and there are $T$ total steps,  $|\hat\beta_N(x)|, |\beta(x)|\leq r_{\max} T$  always. Consequently, $|\hat\beta_N(x)-\beta(x)|\leq 2r_{\max} T$
\item For $N$ trajectories, the number of samples in bin $x$, $N_x$ (i.e the number of trajectories that visit a bin $x$ state), has $N_x\to \infty$ for all $x\in \mathcal{B}$ almost surely, due to the non-zero visit probability assumption. As such, the returns falling in bin $x$ are all independently distributed (as they're from different trajectories by first-visit assumption), and identically distributed according to $R(s), s\sim \pi_{\theta}|f(s)=x$. Thus, as rewards are bounded, SLLN applies to $\hat \beta_N(x)$ and so $\hat \beta_N(x)\to \beta(x)$ almost surely for all $x\in\mathcal{B}$.
\item Moreover, since as mentioned $\hat \beta_N$ and $\beta$ are bounded by $r_{\max} T$ always, the bounded convergence theorem shows that $\mathbb{E}[|\hat \beta_N(x)-\beta(x)|]\to 0$ as $N\to \infty$ for all $x$
\item $\sum_{x\in \mathcal{B}}\rho(x)=\mathbb{E}[\sum_{t=1}^T\mathbbm{1}_{s_t^n \text{ is any state}}]= T$ is finite
\item As $\mathcal{B}$ is countable, it is possible to pick $\mathcal{T}\subset \mathcal{B}$ finite that $\sum_{s\in \mathcal{B}\backslash\mathcal{T}} \rho(s) <\delta$ for any $\delta > 0$ (for instance by enumerating the bins and truncating the sequence per definition of convergence)
\end{itemize}

Thus, pick $\mathcal{T}$ such that $\sum_{s\notin \mathcal{T}} \rho(s) < \frac{\epsilon}{6r_{\max} T}$. As $\mathcal{T}$ is finite, $\sup_{x\in \mathcal{T}} \mathbb{E}[|\hat \beta_x(s)-\beta(s)|]\to 0$ (a finite number of sequences converge uniformly). Thus, there exist $M_\mathcal{T}$ such that for all $N>M_\mathcal{T}$,  have $\sup_{x\in \mathcal{T}} \mathbb{E}[|\hat \beta_N(x)-\beta(x)|]\leq \frac{\epsilon}{3T}$. Then we have for $N>M_\mathcal{T}$:
$$\begin{aligned}\sum_{x\in \mathcal{B}} \rho(x)\mathbb{E}\left[|\hat \beta_N(x)-\beta(x)|\right]&=\sum_{x\in \mathcal{T}} \rho(x)\mathbb{E}\left[|\hat \beta_N(x)-\beta(x)|\right]+\sum_{x\in \mathcal{B}\backslash\mathcal{T}} \rho(x)\mathbb{E}\left[|\hat \beta_N(x)-\beta(x)|\right]\\&\leq\sup_{x\in \mathcal{T}} \mathbb{E}[|\hat \beta_N(x)-\beta(x)|]\sum_{x\in \mathcal{T}} \rho(x)+\sum_{x\in \mathcal{B}\backslash\mathcal{T}} \rho(x)\mathbb{E}\left[2r_{\max} T\right] 
\\&\leq \frac{\epsilon}{3T}\sum_{x\in \mathcal{T}} \rho(x)+2r_{\max} T \sum_{x\in \mathcal{B}\backslash\mathcal{T}} \rho(x)
\\&\leq \frac{\epsilon}{3T}T+2r_{\max} T\frac{\epsilon}{6r_{\max} T}
\\& =\frac{2\epsilon}{3}<\epsilon\end{aligned}$$

This establishes $K_N\to 0$ as $N\to\infty$. It thus follows that 
$$\frac{1}{N}\sum_{n=1}^N \sum_{t=1}^T  \hat A_t^n\nabla_\theta \log \pi_\theta(a_t^n|s_t^n)\to^\mathbb{P} \nabla_\theta\eta(\theta)$$ as required $\square$

\section{Environments}
\label{section:environmentdetails}
\paragraph{Cart Pole} A pole is attached to a cart that moves along a track. The goal is to push the cart left or right to keep the pole upright. The environment features a continuous observation space representing the cart’s (angular) position and velocity, and a discrete action space with left and right movements. Reference hyper-parameters are given in the CleanRL implementation. 

\paragraph{Cliff Walking} The agent must traverse a $4\times 12$ grid world from the bottom-left to the bottom-right corner, avoiding a cliff along the bottom row. The observation space is discrete, representing the agent's grid position, while the action space consists of four directional moves: up, down, left, and right. Reference hyper-parameters are given in the CleanRL implementation. 

\paragraph{Lunar Lander} A lander starts from the top of the environment and must safely land on a designated pad. The agent receives continuous observations of its position, velocity, angle, angular velocity, and leg-ground contact status. The action space is discrete, controlling the main engine and two orientation engines (left and right). Rewards and penalties are provided for successful landings, crashes, proximity to the landing pad, velocity reduction, and engine usage. Reference hyper-parameters are obtained from~\citep{PPO-LunarLander-v2}.

\paragraph{MuJoCo Half Cheetah} MuJoCo (Multi-Joint dynamics with Contact) is a physics-based robotics simulation. The Half Cheetah environment features a cheetah-like robot with two legs and six joints. The agent applies torques to the joints to propel the robot forward as quickly as possible. Both the observation and action spaces are continuous. Reference hyper-parameters are given in the CleanRL implementation. 


\end{document}